\documentclass{article}

\usepackage{arxiv}

\usepackage[utf8]{inputenc} 
\usepackage[T1]{fontenc}    
\usepackage{url}            
\usepackage{booktabs}       
\usepackage{amsmath,amssymb,amsfonts}
\usepackage{nicefrac}       
\usepackage{microtype}      

\usepackage{mathtools}

\usepackage{graphicx}

\usepackage{caption}

\usepackage{subcaption}

\usepackage{amssymb}

\usepackage{algorithmic}

\usepackage{algorithm} 
\usepackage{multirow}
\usepackage[table,xcdraw]{xcolor}

\usepackage[pdfencoding=auto, psdextra]{hyperref}

\usepackage{doi}
\usepackage{eso-pic}

\newtheorem{definition}{Definition}

\usepackage{draftwatermark}

\usepackage{fix-cm}

\SetWatermarkText{Preprint}
\SetWatermarkScale{0.8}
\SetWatermarkLightness{0.9}
\SetWatermarkAngle{45}

\title{Graph Representation Learning of \\ Lightweight IoT Ciphers}

\author{Jonathan Cook*,
        Sabih ur Rehman*
        and M. Arif Khan*\\ 
        {*School of Computing, Mathematics and Engineering, Charles Sturt University, Australia}}

\begin{document}
\maketitle

\begin{abstract}
	SIMON and SIMECK belong to a family of Lightweight Cryptographic Algorithms (LCAs) based on the Feistel block cipher, designed for Internet of Things (IoT) devices. As with all Feistel ciphers, they are susceptible to differential cryptanalysis, necessitating rigorous resilience evaluations. While state-of-the-art techniques leverage heuristics and sampling to improve efficiency, little work has applied Machine Learning (ML) guided Graph Representation Learning (GRL) to efficiently identify and visualise high-probability differential clusters. We address this gap by introducing an efficient feature engineering strategy that extracts four differential attributes from a partial Difference Distribution Table (pDDT), revealing structural information concealed in raw differential data. Utilising the enriched features, we construct and compare three ML-guided directed graphs for SIMON$32$ and SIMECK$32$ using K-Nearest Neighbour (KNN), Decision Trees (DT), and Random Forests (RF). To the best of our knowledge, our framework produces the first graph-based visualisation of the differential clustering effect, in which high-probability single-bit differentials form geometrically close clusters in the learned embedding. All three models achieve a precision of $1.0$ in identifying high-probability differentials, confirming zero false positives. KNN achieves the strongest cluster separation, the highest F1 score and the lowest graph construction time of approximately $2.3$ seconds, while DT and RF produce optimal paths with near-perfect regression. The results are consistent across both LCAs, demonstrating the applicability of the framework to other AND-rotation LCA families.
\end{abstract}


%
%
\textbf{Preprint note:} This is the author’s version of a paper accepted at the 33rd International Conference on Neural Information Processing (ICONIP 2026). The final authenticated version will appear in Springer Communications in Computer and Information Science (CCIS).

\section{Introduction}\label{Sec:Intro}

Internet of Things (IoT) devices face significant security, privacy, and integrity challenges due to their resource-constrained environments \cite{ali2018cyber, cook2023security, meneghello2019iot}. Several Lightweight Cryptographic Algorithms (LCAs) suitable for resource-constrained applications have been developed in recent years, beginning with SIMON and SPECK in $2013$ \cite{beaulieu2015simon}, with SIMON being designed specifically for IoT devices. In 2015, researchers at the University of Waterloo proposed SIMECK, combining properties of both \cite{yang2015simeck}. SIMON and SIMECK employ bitwise-AND rotation operations \cite{stallings2017principles}, which, although efficient, are known to be vulnerable to adversarial attacks, necessitating rigorous security evaluations using cryptanalytic methods \cite{edgar2017science}, ideally targeting as many rounds as possible while maintaining efficiency \cite{de2006introduction}. 

Existing techniques improve cryptanalysis through optimisation strategies, Machine Learning (ML) and graphs, with many using a Difference Distribution Table (DDT) or partial DDT (pDDT) \cite{biryukov2014automatic, cook2024cryptanalysis, cook2024lightweight, cook2026raid, dwivedi2023security, gohr2019improving, so2020deep}. However, a pDDT comprises independent differential records with limited differential information, hindering advances utilising contemporary digital technologies, with efficiency degrading as the table size grows with word size \cite{cook2024lightweight, dwivedi2023security}. To address this, \cite{cook2026raid} proposed efficient FE techniques of pDDT differentials to extract additional features from characteristics identified in \cite{cook2025impact}, aiding ML predictions in a recommendation system. The optimal model exploiting the observed clustering effect of differentials in SIMON \cite{leurent2021clustering} via K-Means clustering. However, the clustering effect in SIMON and SIMECK via graph representations has not yet been investigated. While a pDDT has previously been transformed into a graph structure using Neo4j \cite{cook2024cryptanalysis}, a significant clustering effect was not observed due to the absence of learnable features. By leveraging additional engineered features and Graph Representation Learning (GRL) \cite{celikkanat2022nodesig, kammari2024adaptive}, high-probability differential regions can be identified, leading to the following research question: \textit{How can ML algorithms be utilised in GRL to identify high probability differential clusters, and how do various models compare?}

This paper presents the following contributions. First, we introduce a GRL framework that converts the pDDT into a directed graph in which nodes represent differential pairs and edges encode structural and probabilistic proximity, extending the work of \cite{cook2026raid} and \cite{cook2025impact}. Secondly, to the best of our knowledge, we present the first graph-based visualisation of the differential clustering effect \cite{leurent2021clustering}, demonstrating that high-probability differentials manifest as geometric clusters in a learned feature embedding, transforming an analytically described observation into an algorithmically exploitable structure. Third, we conduct a comparative evaluation of K-Nearest Neighbour (KNN), Decision Trees (DTs), and Random Forests (RFs) as graph construction engines, assessing cluster quality and structural graph qualities. Fourth, we demonstrate that these models yield meaningfully distinct graph topologies, providing insights into the trade-offs between efficiency, graph construction and differential neighbourhood accuracy. The remainder of the document is presented as follows. Section \ref{Sec:Method} describes the methods used in this study. The results of this work are presented in Section \ref{Sec:Results}. We present a formal discussion in Section \ref{Sec:Disc}. Finally, Section \ref{Sec:Concl} presents our concluding comments.

\section{Methodology}\label{Sec:Method}

\begin{figure*}
    \centering
    \includegraphics[width=0.9\linewidth]{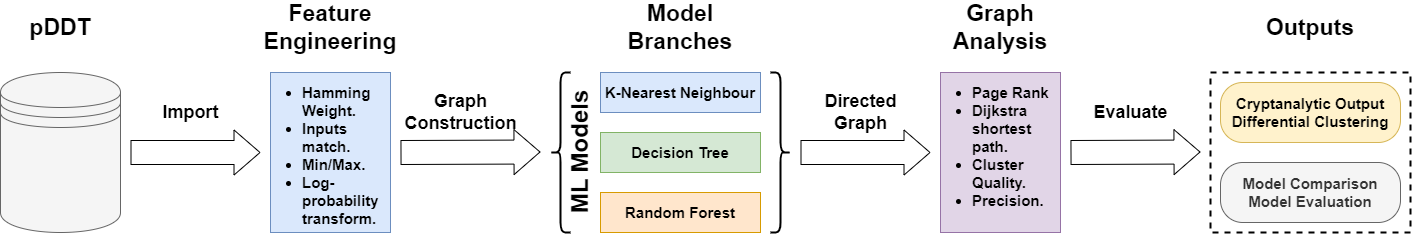}
    \caption{Our methodology}
    \label{fig:contribution}
\end{figure*}

While graphs are typically utilised to describe complex relationships in vast data, dense graphs can complicate interpretation, making them challenging to understand \cite{li2023knowledge}. With a pDDT in place, we utilise ML to guide a KG towards high-probability differentials and the stable state. However, we must first apply Feature Engineer (FE)  to extract additional differential characteristics from the pDDT.
A concise illustration of our methodology is presented in Figure \ref{fig:contribution}.

\subsection{Feature Engineering pDDT differentials}\label{Subsec:PreCompDiff}

To effectively guide ML in identifying high-probability differentials, we augment the pDDT with additional features. The pDDT file is imported into the Python Pandas package, producing a dataframe with null-initialised feature columns to be populated during FE.

\subsubsection{Hamming Weight (HW) of \texorpdfstring{$\Delta_X$}{\Delta\_X} (\texorpdfstring{$\Delta_{X_w}$}{\Delta\_{X\_w}}), \texorpdfstring{$\Delta_Y$}{\Delta\_Y} (\texorpdfstring{$\Delta_{Y_w}$}{\Delta\_{Y\_w}}) and \texorpdfstring{$\Delta_Z$}{\Delta\_Z} (\texorpdfstring{$\Delta_{Z_w}$}{\Delta\_{Z\_w}})}
Counts the number of bits set to one in the binary string of both input differentials, is used to identify single-bit (power-of-two) differentials and the min/max values. 

\subsubsection{Inputs match (\texorpdfstring{$\hat{\Delta}$}{\hat{\Delta}})}
Determines if both input differentials share identical bit positions (e.g., $\Delta_X = \Delta_Y$). A Boolean value is returned as a differential feature.

\subsubsection{Probability transformation}
Transforms the probability score to a logarithmic scale, enabling a scalable representation of probabilities across low threshold configurations.

\subsubsection{\texorpdfstring{$arg~min~max(\Delta_X)$}{arg min max(\Delta\_X)} and \texorpdfstring{$arg~min~max(\Delta_Y)$}{arg min max(\Delta\_Y)}}
Used to identify if input differentials are minima or maxima values in the pDDT word size space, returning a Boolean value derived from the HW.

\subsection{Graph Creation}\label{Subsec:GrCreate}

For our analysis, we compare three ML models, KNN, DT and RF, in guiding a graph construction towards high-probability, non-trivial differentials. 
Although k-means has previously been utilised to improve efficiency and increase the number of rounds targeted, KNN and DT offered efficiency advantages with natural branching to nearby data points \cite{cook2026raid}. The efficiency of KNN and DT underscores their potential to guide a graph towards high-probability differentials. 
Graph size is limited to 500 nodes by importance, producing a transparent structure in which relationships between nodes are readily interpretable.
To evaluate each model, we employ the following definitions:

\begin{definition} \label{Def:GraphExTime}
    Let the experimental time recordings be a set of $T_E = \{t_1, t_2, t_3 \dots t_n\}$ recorded in seconds. The total time ($T$) is the sum of all time recordings of the set $T_E$.  From the set of time recordings, we derive the mean graph execution time ($\bar{T}_{G}$) as $\bar{T}_{G} = \frac{T}{\mid T_E \mid}$.
\end{definition}

\begin{definition}
    The total number of edges in a graph highlights the overall potential attack paths.

\end{definition}

\begin{definition}
    Total number of critical attack paths. Critical attack paths are edges connecting the highest probability differentials.
\end{definition}

\begin{definition}
    Hops from random differential to stable-state differential via edge traversal ($E_{hop}$). Many optimisation strategies employ heuristic search methods beginning at a random differential and continually sampling until a high-probability differential is identified. Employing a random starting differential, the number of hops from the random starting differential to the stable-state node.
\end{definition}

\begin{definition}
    Time per random hop ($T_{hop}$) evaluates the time required to create the graph and the number of hops from a random starting node and is defined as $T_{hop} = \frac{\bar{T}_{G}}{E_{hop}}$.
\end{definition}

Graphs are rendered using a seeded Fruchterman-Reingold force-directed layout \cite{syafiq2026fr}, where node coordinates are initialised by layer towards concentric arrangement. Nodes and edges are coloured and scaled separately to highlight key structural features and facilitate the identification of shortest paths to the stable-state root node.
Graph construction utilises $5$ complementary distance and penalty concepts:

\begin{enumerate}
    \item Euclidean Distance (ED) is a measure that drives the neighbour (KNN) or branch (DT) selection process as a distance to the stable state node and is used in edge weight calculations.
    \item Importance score ($I_s$) is a custom heuristic that ranks nodes by their differential properties, prioritising $\Delta_X = \Delta_Y$, HW $= 1$, and a high Differential Probability (DP).
    \item PageRank (PR) \cite{page1999pagerank} is utilised as a centrality measure, ranking nodes by the significance of their connectivity within the graph.  
    PR scores are boosted post-computation for the stable-state node and high-priority differential types to reflect their cryptanalytic significance.
    \item Hamming Distance (HD) measure on $\Delta_X$ to $\Delta_X'$ and $\Delta_Y$ to $\Delta_Y'$ evaluates single-bit transitions between differentials.
    \item Shortest path from each node to the root node, computed using Dijkstra's Algorithm with edge cost ($cost(v' \rightarrow v) = 1/weight$), where $(v, v')$ is the direct link from node $v$ to node $v'$, quantifying proximity to the root.
\end{enumerate}

\subsubsection{K-Nearest Neighbour}
The KNN-guided graph utilises the HW of $\Delta_X$, $\Delta_Y$, $\Delta_Z$, and the logarithmic probability as features in the feature space. The stable-state root node is defined as the differential with properties of $\Delta_{X_w} = 0, \Delta_{Y_w} = 0$ and $\Delta_{Z_w} = 0$ and the highest probability. If absent, the fallback is $\hat\Delta = 1$ with minimum HW and the highest probability. Node layers are assigned as:
\begin{equation}
    layer = \begin{cases}
                \Delta_{X_w} & \text{if} \hat{\Delta} = 1 \\
                \min(\Delta_{X_w}, \Delta_{Y_w}) + 1 & \textbf{otherwise}.
            \end{cases}
\end{equation}

For each node, the nearest neighbours are enumerated, and directed edges are created towards neighbours with shorter ED to the root. The directed edge weight ($W_e$) is defined as:
\begin{equation}
    W_e = Sim \times P_t,
\end{equation}

\indent where $P_t$ is the target probability and $Sim$ is defined as:

\begin{equation}
    Sim = \frac{1}{1+ED}.
\end{equation}

During rendering, a random non-root node is selected as a starting position, and the shortest path to the stable-state root is visualised, with red edges denoting the optimal traversal path.

\subsubsection{Decision Tree}

The DT-guided graph utilises the same feature space as the KNN, training a Classification And Regression Tree (CART) regressor \cite{bittencourt2004feature} on a blended target:
\begin{equation}
    y = 0.6 \cdot DP+0.4 \cdot S,
\end{equation}

\indent where $S$ is the structural score as: $\hat{\Delta} (0.44)$, power of two of both inputs ($0.24$), inverse total bit count ($0.2$), and the minimum or maximum status ($0.12$). Each node receives a predicted score and leaf ID. Edges are created in two tiers: within-leaf edges connect to higher-scoring neighbours within the same partition, and up-tree edges connect to higher-ranked leaves with a higher mean tree score, consistently directing traversal towards better-scoring nodes. Edge weights combine tree score, structural features, and a Proximity Factor (PF), which penalises long-range edges across the graph. PF is defined as:

\begin{equation}
    PF = \frac{1}{1+ED}.
\end{equation}

\subsubsection{Random Forest}
The RF-guided graph uses the same feature space and blended target as the DT model. However, rather than a single tree, RF aggregates the predictions of independent DTs, each trained on a different sample, with the final score taken as the mean prediction across all estimators.

\begin{equation}
    \hat{y}_{RF} = \frac{1}{N_e}\sum_{t=1}^{N_e}\hat{y}_t,
\end{equation}

\indent where $N_e$ is the number of estimators and $\hat{y}_t$ is the prediction of the $t^{th}$ tree. A key distinction of the RF model is the ensemble variance tracking, where the per-node prediction variance across all estimators is recorded as a confidence measure as:

\begin{equation}
    \sigma^2_i = \text{Var}(\hat{y}_{1, i}, \hat{y}_{2, i}, \dots, \hat{y}_{N_e, i}), \quad C_i = \frac{1}{1+\sigma^2_i},
\end{equation}

\indent where $C_i$ is the confidence score for node $i$, with higher values indicating greater estimator agreement. Leaf assignment for graph edge construction is derived from the first estimator, and the same two-tier within-leaf and up-tree edge creation strategy as DT is applied.

\section{Results}\label{Sec:Results}
In previous sections, we described our feature engineering for differentials and the implementation of ML-guided graphs for identifying high-probability differentials. In this section, we present the results of our work, highlighting efficiency gains, differential feature enhancements and the interconnected relationships of differentials. Through a comparative analysis of 3 ML models, we highlight their key advantages and potential limitations.

\subsection{Graph Analysis}\label{Subsec:GrAn}

We now provide a comparative analysis of the ML-guided graphs of SIMON and SIMECK, which will help us identify key differences in how models process and represent the pDDT data. The graphs consist of multiple layers of vertices, with the stable-state differential centred and highlighted in yellow. Orange and blue vertices in the inner ring denote high-probability differentials with specific bitwise transition states, such as a Hamming weight of $1$ and matching input differences. The graph consists of three edges of different colours. Light grey edges represent background connectivity, typically with longer, suboptimal traversals.
The red edges emphasise the critical guiding paths to the stable-state node. The blue edge highlights the shortest path from a randomly selected starting node to the central stable-state node.

\begin{figure}[htbp]
    \centering
    \begin{subfigure}{0.45\columnwidth}
        \includegraphics[width=\textwidth]{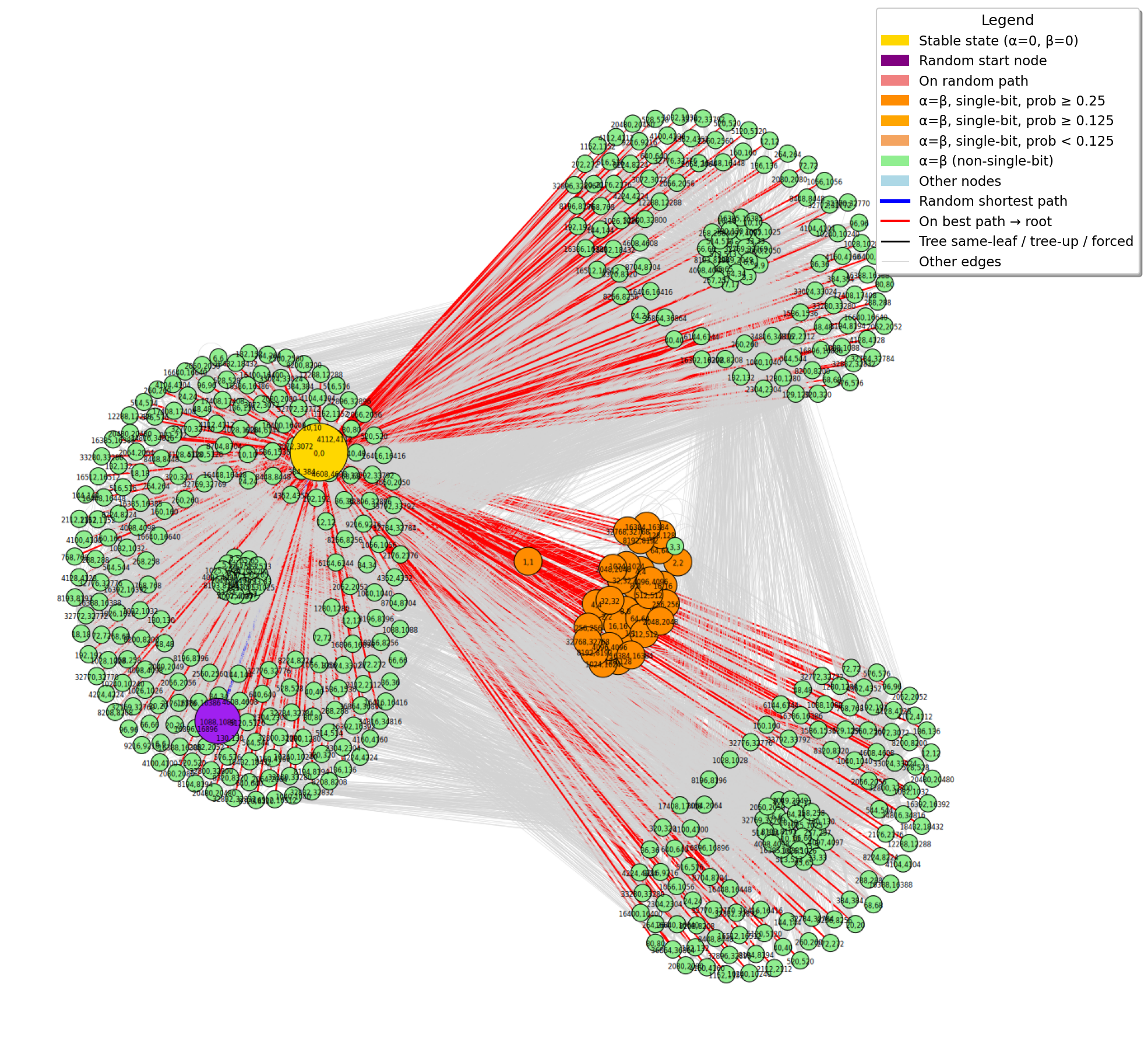}
        \caption{KNN guided graph of SIMON$32$}
        \label{fig:SIMON_knn_guided_wsddt_graph_500_nodes_16bit_65536window_0.1thresh}
    \end{subfigure}
    \hfill
    \begin{subfigure}{0.45\columnwidth}
        \includegraphics[width=\textwidth]{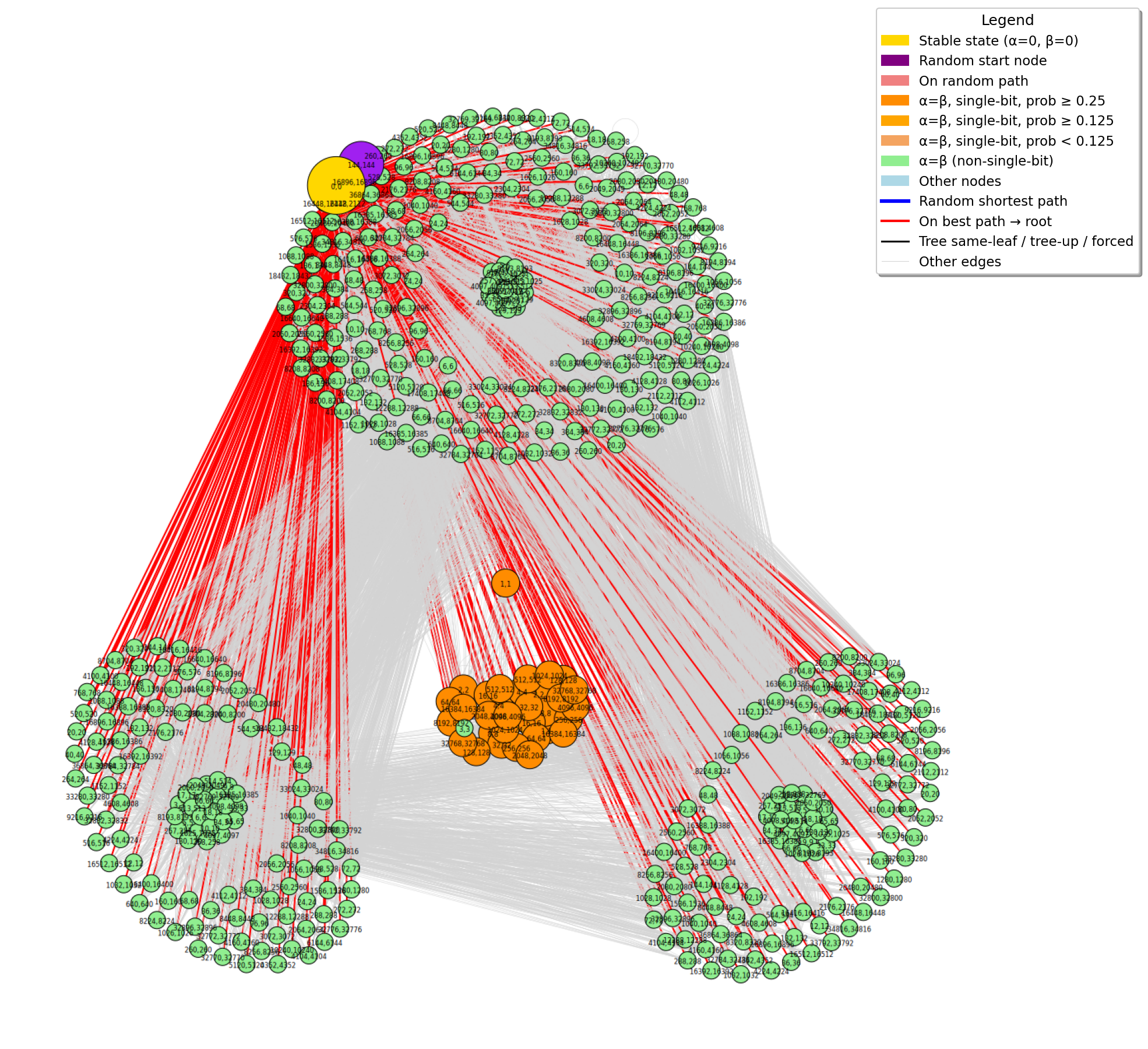}
        \caption{KNN guided graph of SIMECK$32$}
        \label{fig:SIMECK_knn_guided_wsddt_graph_150_nodes_16bit_8window_0.1thresh}
    \end{subfigure}
    \caption{KNN guided graphs.}
    \label{fig:simon_simeck_knn}
\end{figure}

We begin our graph analysis of the KNN guided graph of SIMON32 and SIMECK32, presented in Figures \ref{fig:SIMON_knn_guided_wsddt_graph_500_nodes_16bit_65536window_0.1thresh} and \ref{fig:SIMECK_knn_guided_wsddt_graph_150_nodes_16bit_8window_0.1thresh} respectively. Both graphs render 500 nodes under an identical threshold configuration, enabling direct structural comparison between the two LCAs. The most prominent observation across both KNN figures is the tight spatial clustering of orange nodes, which are differentials that satisfy both $\hat{\Delta} = 1$ and $\Delta_{X_w} = 1$, corresponding to the highest-probability single-bit differentials. In Figure \ref{fig:SIMECK_knn_guided_wsddt_graph_150_nodes_16bit_8window_0.1thresh}, this cluster forms a dense, geometrically cohesive group at the centre of the graph. In Figure \ref{fig:SIMON_knn_guided_wsddt_graph_500_nodes_16bit_65536window_0.1thresh}, the same cluster occupies the upper-central region and exhibits a slightly greater dispersion, reflecting the broader spread of high probability differentials in SIMON relative to SIMECK. This geometric separation of high-probability differentials from broader node populations provides graph-based visual confirmation of the differential clustering effect \cite{leurent2021clustering}.

Beyond the orange cluster, both graphs exhibit a structure in which green nodes form broad peripheral groupings. In both graphs, three distinct lobes are visible. However, SIMECK exhibits tighter cohesion between the two outer lobes than SIMON. These structural contrasts reflect the distinct design differences of the LCAs, with SIMON's wider lobe distribution suggesting a less centralised differential dependency. Additionally, the stable-state root nodes occupy different positions in both graphs. In SIMECK, the root node is anchored at the top centre, with red paths fanning down across the graph. In SIMON, the root node settles in the upper-right centre, with red paths fanning outward toward prominent differentials. In both figures, KNN consistently produces paths to the root node, regardless of the random node's starting position. The overall density of the grey edges reflects the lazy learning nature of KNN, in which each node retains connections to multiple neighbours rather than committing to a single branch. While this technique produces a dense graph, the structural hierarchy remains interpretable, with the red critical paths being clearly distinguished from the broader connectivity.

\begin{figure}[htbp]
    \centering
    \begin{subfigure}{0.45\columnwidth}
        \includegraphics[width=\textwidth]{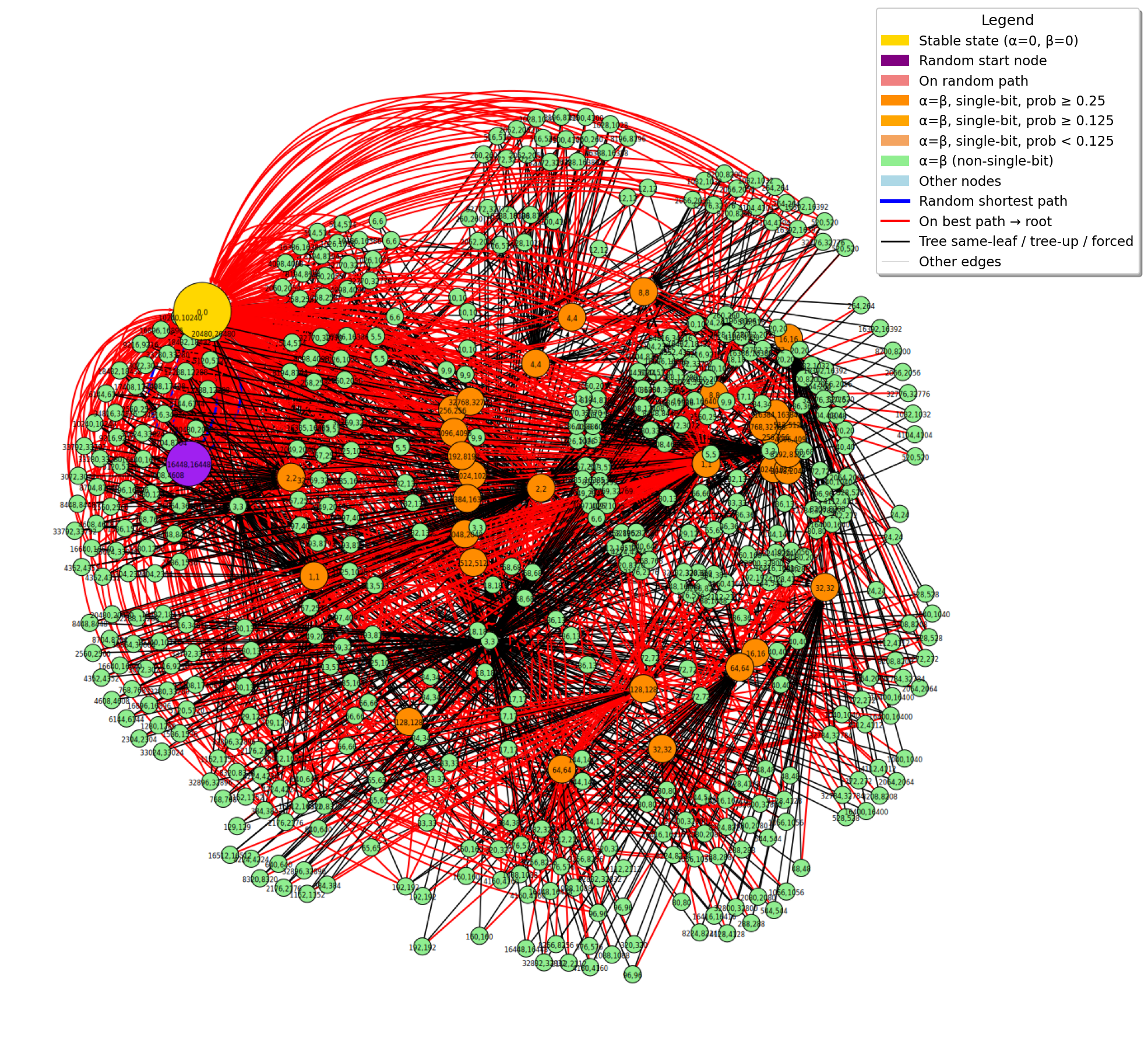}
        \caption{DT guided graph of SIMON$32$}
        \label{fig:SIMON_dt_guided_wsddt_graph_500_nodes_16bit_65536window_0.1thresh}
    \end{subfigure}
    \hfill
    \begin{subfigure}{0.45\columnwidth}
        \includegraphics[width=\textwidth]{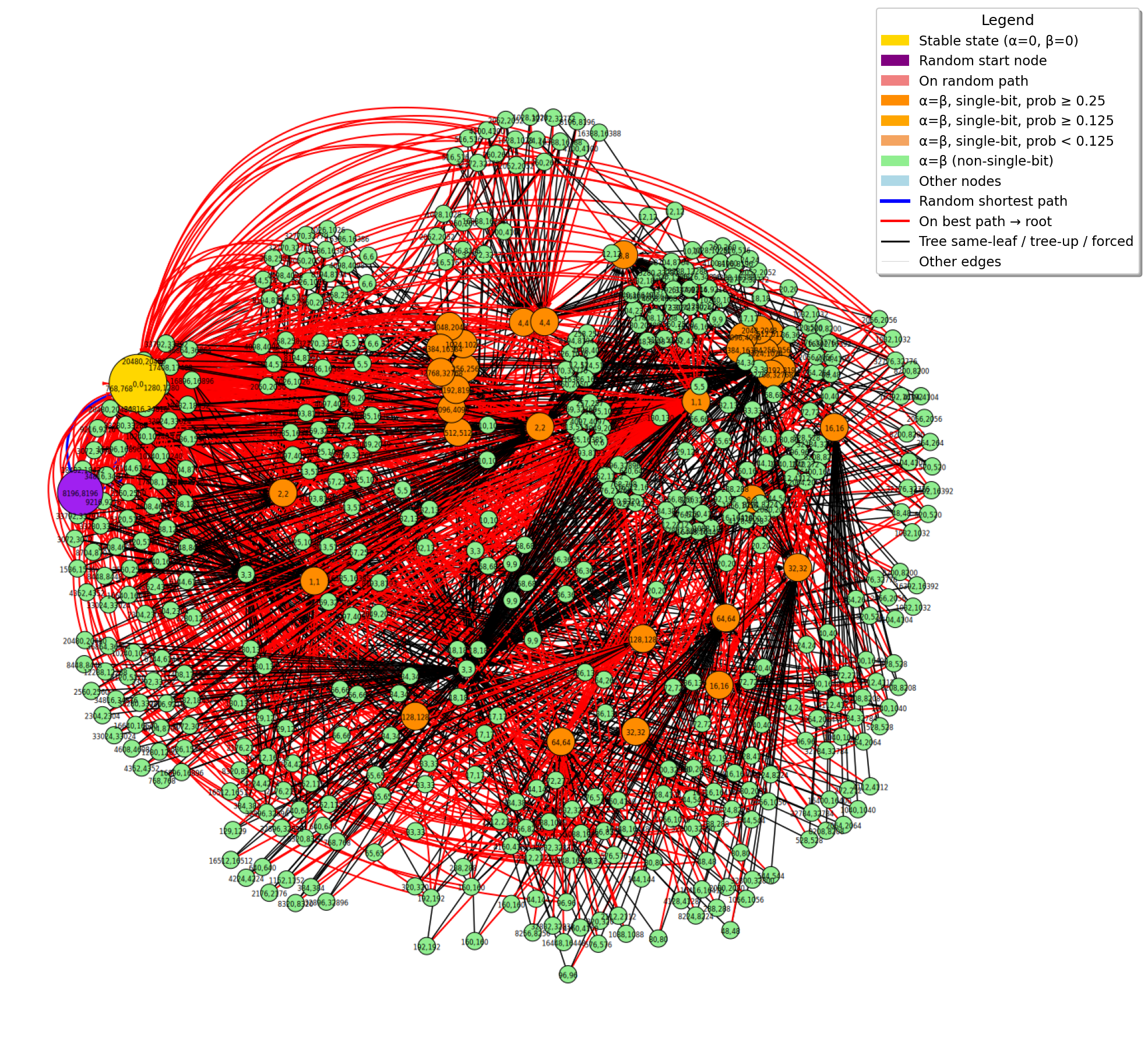}
        \caption{DT guided graph of SIMECK$32$}
        \label{fig:SIMECK_dt_guided_wsddt_graph_500_nodes_16bit_65536window_0.1thresh}
    \end{subfigure}
    \caption{DT guided graphs.}
    \label{fig:simon_simeck_dt}
\end{figure}

Figures \ref{fig:SIMON_dt_guided_wsddt_graph_500_nodes_16bit_65536window_0.1thresh} and \ref{fig:SIMECK_dt_guided_wsddt_graph_500_nodes_16bit_65536window_0.1thresh} present the DT-guided graphs for SIMON32 and SIMECK32, respectively. The structural contrast with KNN is immediately apparent. Where KNN produces a fan-shaped topology with a tight peripheral orange cluster and distinct green lobes, the DT guided graph exhibits a dense, approximately circular arrangement in which nodes are distributed more uniformly across the graph interior. The most notable structural difference concerns the placement of orange single-bit differentials. Rather than forming isolated clusters, as in KNN, high-probability nodes with $\hat{\Delta} = 1$ and $\Delta_{X_w} = 1$ are distributed throughout the central region, intermixed with the broader node population. This reflects the two-tier edge construction of the DT model, in which within-leaf edges bind structurally similar nodes locally and up-tree edges connect across partitions, pulling high-scoring differentials inward rather than allowing them to aggregate at a spatial periphery. Additionally, the black edges are clearly visible in both figures, a feature absent in the KNN graphs, and provide evidence that the leaf-based partitioning guides the connectivity. The density of red critical-path edges is notably higher in both graphs relative to their KNN counterparts. This indicates that more nodes lie on paths to the root node. The root node is positioned in the left centre of both graphs, with the random starting node located at the far left, near the root, with a recoverable path in both instances. Both DT graphs are also structurally more similar to one another than under KNN guidance, suggesting that the CART regressor's blended probability-structure target produces a consistent partitioning topology across the two LCAs.

\begin{figure}[htbp]
    \centering
    \begin{subfigure}{0.45\columnwidth}
        \includegraphics[width=\textwidth]{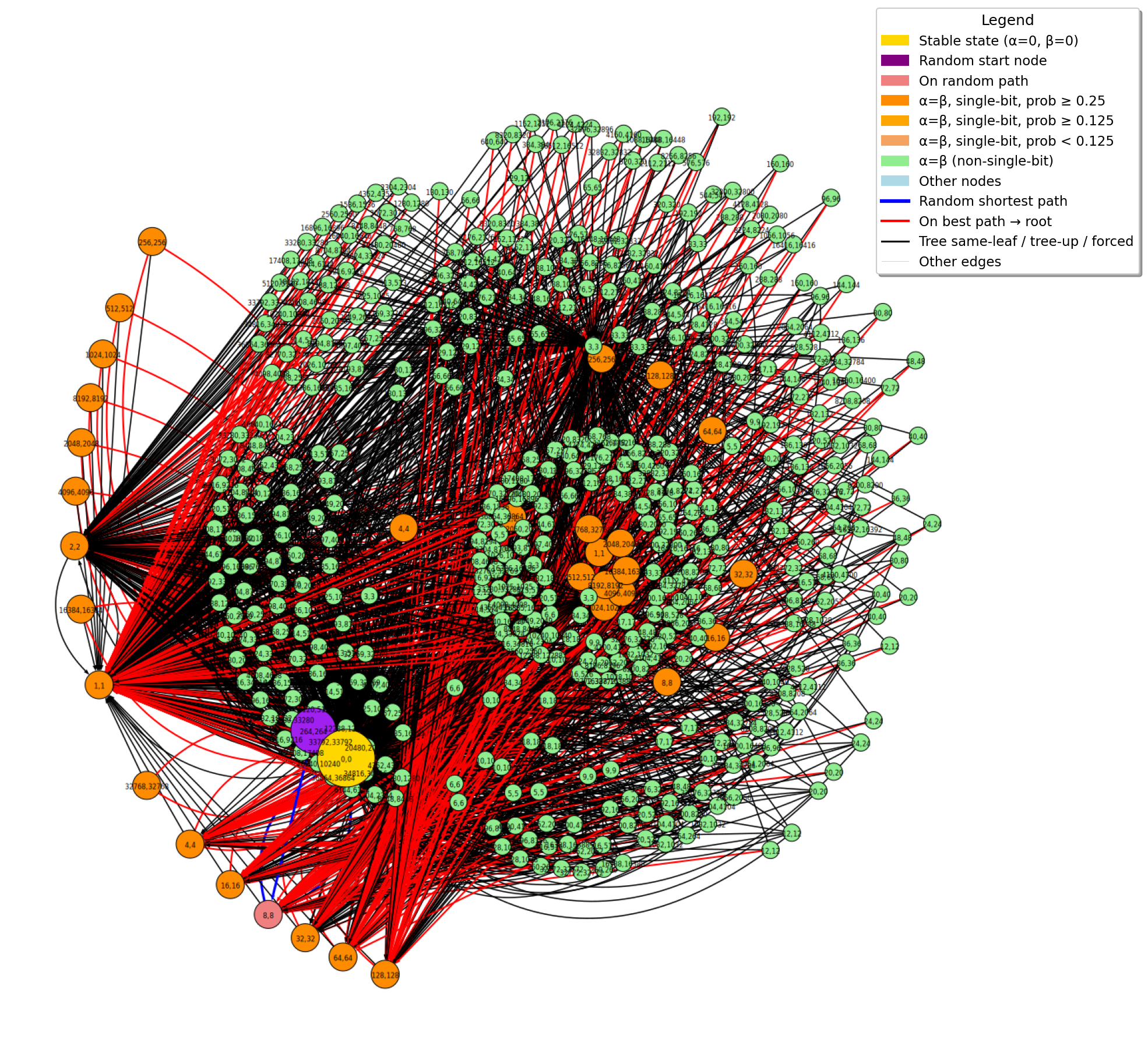}
        \caption{RF guided graph of SIMON$32$}
        \label{fig:SIMON_rf_guided_wsddt_graph_500_nodes_16bit_65536window_0.1thresh}
    \end{subfigure}
    \hfill
    \begin{subfigure}{0.45\columnwidth}
        \includegraphics[width=\textwidth]{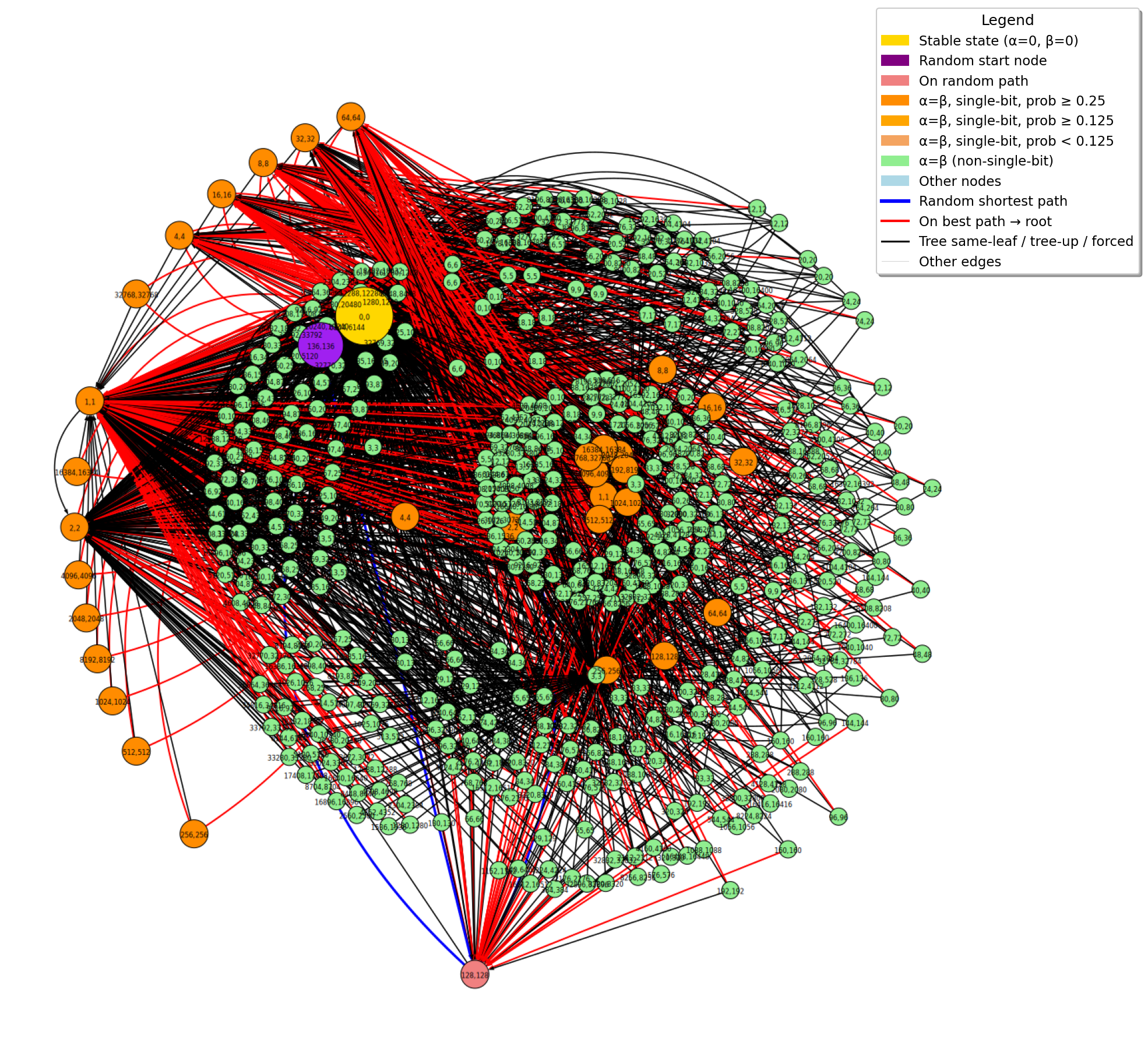}
        \caption{RF guided graph of SIMECK$32$}
        \label{fig:SIMECK_rf_guided_wsddt_graph_500_nodes_16bit_65536window_0.1thresh}
    \end{subfigure}
    \caption{RF guided graphs.}
    \label{fig:simon_simeck_rf}
\end{figure}

Figures \ref{fig:SIMON_rf_guided_wsddt_graph_500_nodes_16bit_65536window_0.1thresh} and \ref{fig:SIMECK_rf_guided_wsddt_graph_500_nodes_16bit_65536window_0.1thresh} present the RF-guided graphs for SIMON32 and SIMECK32, respectively. The RF graphs exhibit a topology distinct from both KNN and DT, with several structurally notable features. The most visually prominent characteristic is the emergence of orange single-bit differentials along the left boundary of both graphs, which is spatially separated from the broader node population. This peripheral positioning reflects the ensemble-averaging effect of RF, in which the model consistently assigns high scores to the most structurally favoured differentials, thereby pulling them outward as high-centrality anchor points. Red critical-path edges concentrate heavily around these left-spine nodes, indicating they serve as primary intermediaries on paths to the root.

A secondary orange cluster is visible in the central region of both graphs, interspersed with the green node population, as in the DT graphs. However, a further distinctive feature is the presence of a lone high-probability orange node at the lower periphery in SIMECK, isolated from the main body by several red convergence edges, suggesting that the RF model has assigned this node a particularly high importance whilst placing it outside the primary leaf partitions. Black edges are more dominant in the RF graphs than in DT, reflecting the denser within-leaf and up-tree scaffolding that results from leaf assignments derived from the ensemble scores. The green lower-probability nodes exhibit a more structured, concentric ring pattern around the graph interior than is observed in DT, indicative of the smoother score gradients produced by ensemble averaging. Overall, red critical paths are fewer in number but more spatially localised than in the DT graphs, concentrated on the orange spine and central cluster rather than uniformly distributed, suggesting that RF produces a sparser yet more targeted set of high-confidence differential pathways to the root node.

While the above figures provide insights into the graph's structure, additional information can be extracted from the corresponding data. As shown, KNN is significantly more efficient than both DT and RF at guiding a graph to the stable state. However, as stated above, it does so by exploring many more paths, which may not be as optimal as the paths identified by DT and RF. Indeed, examining the total number of relationships reveals a more exhaustive search by KNN, in which all nodes are considered potential neighbours. Across all relationships, DT and RF show far fewer optimal paths, demonstrating the rigidity of the decision-making process. While the number of optimal paths of DT and RF is significantly lower than KNN, all models illustrate similar reductions of $\geq 90\%$ from the total number of relationships. However, a notable feature is the model's ability to identify a path from a random node to the stable state. As shown, KNN reliably identified a path from a random node to the stable state, while both DT and RF fail with smaller samples of the pDDT data. Although the paths identified by KNN typically include more steps over a greater distance, its lazy learning approach facilitates robust pathfinding, reliably guiding a path from a random node to the stable state.

\subsection{Model Evaluation}

\begin{figure}
     \centering
     \begin{subfigure}[b]{0.48\textwidth}
         \centering
         \includegraphics[width=\textwidth]{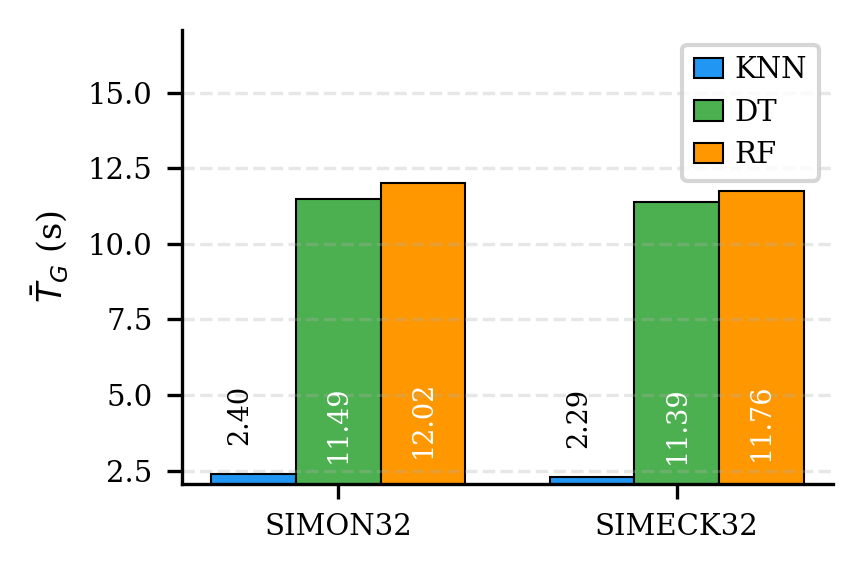}
         \caption{Graph creation time}
         \label{fig:graph_creation_time}
     \end{subfigure}
     \hfill
     \begin{subfigure}[b]{0.48\textwidth}
         \centering
         \includegraphics[width=\textwidth]{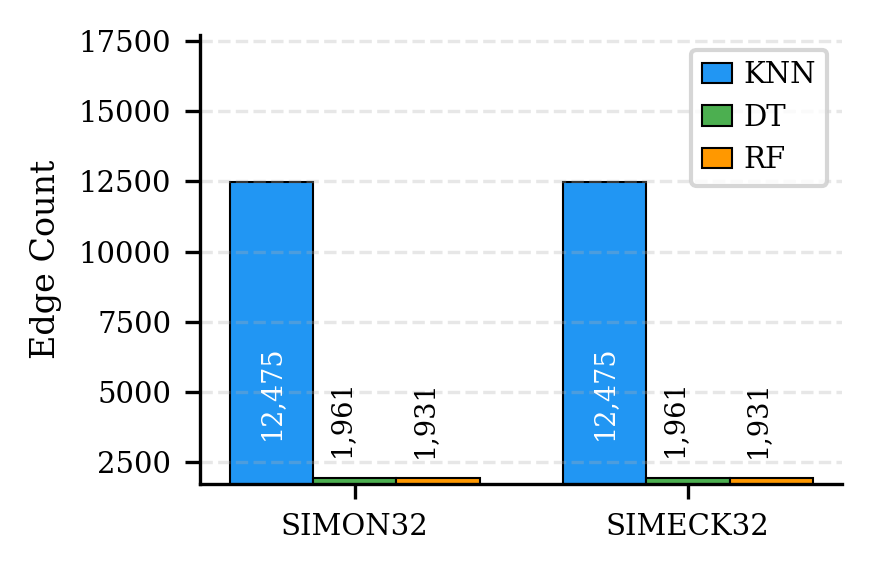}
         \caption{Total relationships}
         \label{fig:total_relationships}
     \end{subfigure}
     \hfill
     
    \caption{Model comparison for SIMON$32$ and SIMECK$32$}
    \label{fig:model_comparison_grid}
\end{figure}

While the above figures in Section \ref{Subsec:GrAn} provide insights into the structure of the graph, additional information can be extracted from the corresponding data. Table \ref{tab:construction_regression} presents the construction metrics for all three models and LCA configurations. An initial observation is that metrics are identical within each model type across both LCAs, confirming that the graph construction algorithms respond deterministically to the shared pDDT structural properties of SIMON$32$ and SIMECK$32$ rather than to algorithm-specific designs. As shown in Figure \ref{fig:graph_creation_time}, KNN constructs graphs approximately five times faster than DT or RF. Figure \ref{fig:total_relationships} illustrates that KNN produces $12475$ edges compared to approximately $1931$ and $1961$ for DT and RF models, respectively. This increase in edge density reflects KNN's lazy learning strategy, in which all $K$ nearest neighbours of each node are retained as potential edges, whereas DT and RF commit only the within-leaf and up-tree connections mandated by the CART partition structure.

\begin{table}
\caption{Graph construction and regression metrics}
\label{tab:construction_regression}
\centering
\tiny
\begin{tabular}{|l|l|r|r|r|r|r|}
\hline
\rowcolor[HTML]{000000}
{\color{white}\textbf{LCA}} &
{\color{white}\textbf{Model}} &
{\color{white}\textbf{Time (s)}} &
{\color{white}\textbf{Edges}} &
{\color{white}\textbf{R\textsuperscript{2}}} &
{\color{white}\textbf{MAE}} &
{\color{white}\textbf{RMSE}} \\
\hline
\multirow{3}{*}{SIMON$32$}
  & KNN & 2.40  & 12{,}475 & $-6.6182^*$ & 0.1438 & 0.1522 \\
  & DT  & 11.49 & 1{,}961  & 0.9974            & 0.0003 & 0.0040 \\
  & RF  & 12.02 & 1{,}931  & 0.9964            & 0.0002 & 0.0047 \\
\hline
\multirow{3}{*}{SIMECK$32$}
  & KNN & 2.29  & 12{,}475 & $-6.6182^*$ & 0.1438 & 0.1522 \\
  & DT  & 11.39 & 1{,}961  & 0.9974            & 0.0003 & 0.0040 \\
  & RF  & 11.76 & 1{,}931  & 0.9964            & 0.0002 & 0.0047 \\
\hline
\end{tabular}
\end{table}

\begin{figure}[t]
    \centering
    \includegraphics[width=0.5\linewidth]{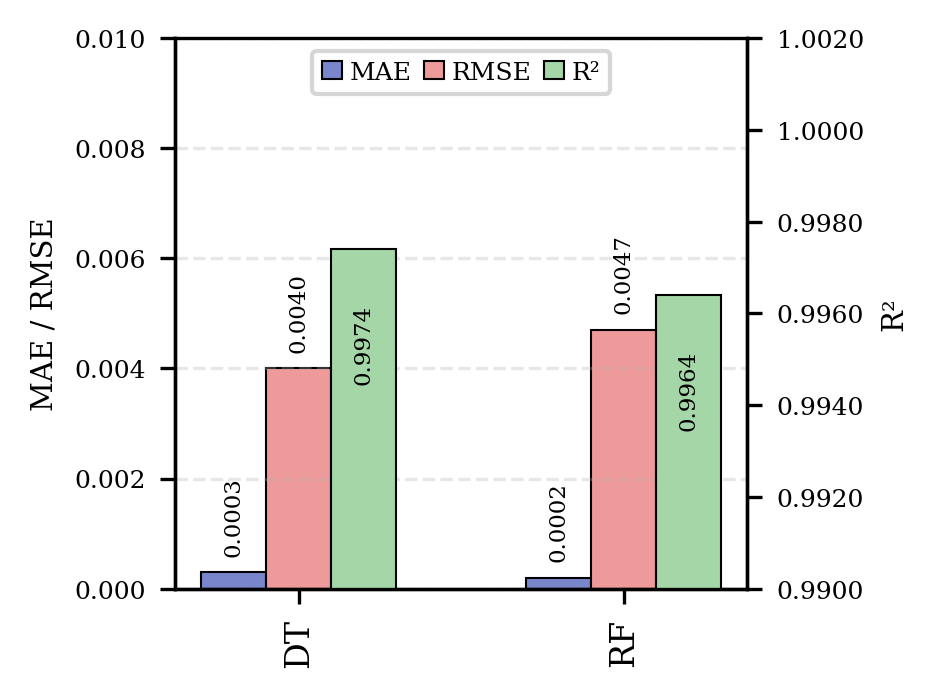}
        \caption{Regression metrics for DT and RF.}
        \label{fig:regression_metrics}
\end{figure}

For DT and RF, regression performance is near-perfect, with $R^2 = 0.9974$ and $0.9964$ respectively, Mean Absolute Error (MAE) below $0.0004$ and Root Mean Square Error (RMSE) below $0.005$, as shown in Figure \ref{fig:regression_metrics}. The marginally higher RMSE of RF relative to DT ($0.0047$ versus $0.0040$) reflects the smoothing effect of ensemble averaging on the decision boundaries of individual trees. KNN reports a strongly negative $R^2$ because its surrogate score, inverted distance-to-root in the feature embedding,  is not a regression prediction in the supervised sense and exhibits high variance relative to the probability distribution. This value is therefore not comparable with the DT and RF results.

\subsubsection{Classification Metrics}

\begin{figure}
    \centering
    \includegraphics[width=0.5\textwidth]{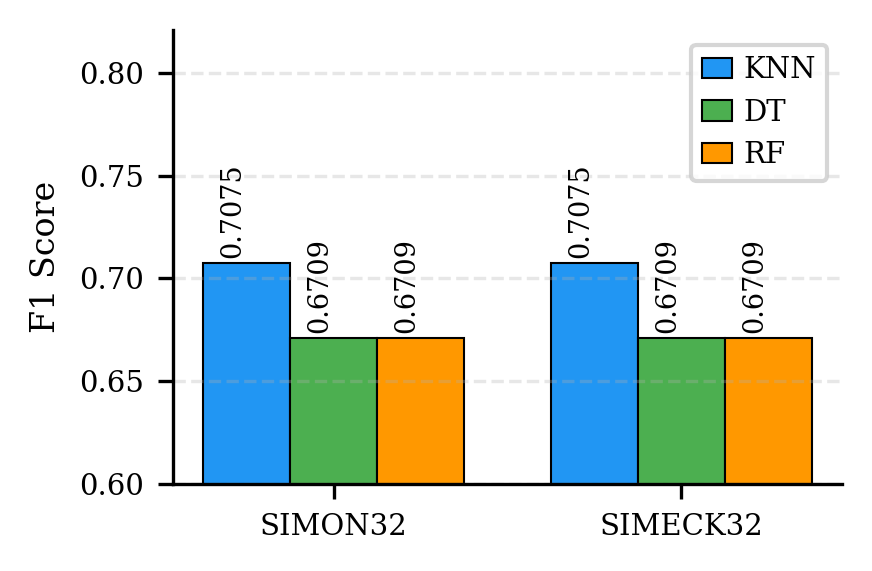}
         \caption{F1 score}
         \label{fig:f1_score}
\end{figure}

Table \ref{tab:classification} presents supplementary classification metrics for the task of identifying single-bit, matching input differentials. A precision of $1.0000$ across all models and both LCAs is the most significant result presented. No model produces a false positive, meaning that whenever a node is classified as belonging to the highest-priority single-bit group, that classification is always correct. This is cryptographically meaningful, as all three graphs correctly identify high-priority candidate differentials for attack. As shown in Figure \ref{fig:f1_score}, KNN achieves a higher F1 score than DT and RF, driven by higher recall, indicating that KNN's dense neighbourhood structure recovers a larger proportion of the high-priority differentials within the graph's top-ranked nodes. DT and RF yield identical classification results, consistent with their shared target function and equivalent leaf-based node prioritisation.

\begin{table}
\caption{Classification metrics for $\hat{\Delta}=1$, $\Delta_{X_w}=1$ differential identification}
\label{tab:classification}
\centering
\tiny
\begin{tabular}{|l|l|r|r|r|r|}
\hline
\rowcolor[HTML]{000000}
{\color{white}\textbf{LCA}} &
{\color{white}\textbf{Model}} &
{\color{white}\textbf{Accuracy}} &
{\color{white}\textbf{Precision}} &
{\color{white}\textbf{Recall}} &
{\color{white}\textbf{F1}} \\
\hline
\multirow{3}{*}{SIMON$32$}
  & KNN & 0.5474 & 1.0000 & 0.5474 & 0.7075 \\
  & DT  & 0.5048 & 1.0000 & 0.5048 & 0.6709 \\
  & RF  & 0.5048 & 1.0000 & 0.5048 & 0.6709 \\
\hline
\multirow{3}{*}{SIMECK$32$}
  & KNN & 0.5474 & 1.0000 & 0.5474 & 0.7075 \\
  & DT  & 0.5048 & 1.0000 & 0.5048 & 0.6709 \\
  & RF  & 0.5048 & 1.0000 & 0.5048 & 0.6709 \\
\hline
\end{tabular}
\end{table}

\subsubsection{Clustering Metrics}

Table \ref{tab:clustering} presents the primary cryptographic evaluation metric, which is the mean Dijkstra distance from each node group to the stable-state root, and the derived Cluster Quality (CQ) score. The CQ score quantifies the degree to which the graph separates high-probability single-bit differentials from the broader node population. A higher CQ score indicates that such nodes sit substantially closer to the root than the remaining nodes, providing a graph-structural representation of the differential clustering effect observed in \cite{leurent2021clustering}.

\begin{table}
\caption{Clustering proximity metrics }
\label{tab:clustering}
\centering
\tiny
\begin{tabular}{|l|l|r|r|r|r|}
\hline
\rowcolor[HTML]{000000} 
{\color[HTML]{FFFFFF} \textbf{Model}} & {\color[HTML]{FFFFFF} \textbf{Node Group}}        & {\color[HTML]{FFFFFF} \textbf{Count}} & {\color[HTML]{FFFFFF} \textbf{Mean dist.}} & {\color[HTML]{FFFFFF} \textbf{Std dist.}} & {\color[HTML]{FFFFFF} \textbf{CQ Score}} \\ \hline
                                      & Root ($\alpha=0$, $\beta=0$)                      & 1                                     & 0.0000                                     & 0.0000                                     &                                          \\
                                      & $\hat{\Delta}=1$, $\Delta_{X_w}=1$, $p \geq 0.25$ & 32                                    & 3.5431                                     & 0.1269                                     &                                          \\
                                      & $\hat{\Delta}=1$ (non-single-bit)                 & 507                                   & 7.2090                                     & 0.8017                                     &                                          \\
\multirow{-4}{*}{KNN}                 & Other nodes                                       & 4{,}460                               & 6.8103                                     & 1.1776                                     & \multirow{-4}{*}{1.9221}                 \\ \hline
                                      & Root ($\alpha=0$, $\beta=0$)                      & 1                                     & 0.0000                                     & 0.0000                                     &                                          \\
                                      & $\hat{\Delta}=1$, $\Delta_{X_w}=1$, $p \geq 0.25$ & 32                                    & 1.5731                                     & 0.1589                                     &                                          \\
                                      & $\hat{\Delta}=1$ (non-single-bit)                 & 507                                   & 3.1519                                     & 0.4102                                     &                                          \\
\multirow{-4}{*}{DT}                  & Other nodes                                       & 4{,}460                               & 2.4305                                     & 0.5999                                     & \multirow{-4}{*}{1.5450}                 \\ \hline
                                      & Root ($\alpha=0$, $\beta=0$)                      & 1                                     & 0.0000                                     & 0.0000                                     &                                          \\
                                      & $\hat{\Delta}=1$, $\Delta_{X_w}=1$, $p \geq 0.25$ & 32                                    & 1.5731                                     & 0.1589                                     &                                          \\
                                      & $\hat{\Delta}=1$ (non-single-bit)                 & 507                                   & 3.1519                                     & 0.4102                                     &                                          \\
\multirow{-4}{*}{RF}                  & Other nodes                                       & 4{,}460                               & 2.4305                                     & 0.5999                                     & \multirow{-4}{*}{1.5450}                 \\ \hline
\end{tabular}
\end{table}

\begin{figure}[ht]
    \centering
    \begin{subfigure}[t]{0.48\columnwidth}
        \centering
        \includegraphics[width=\linewidth]{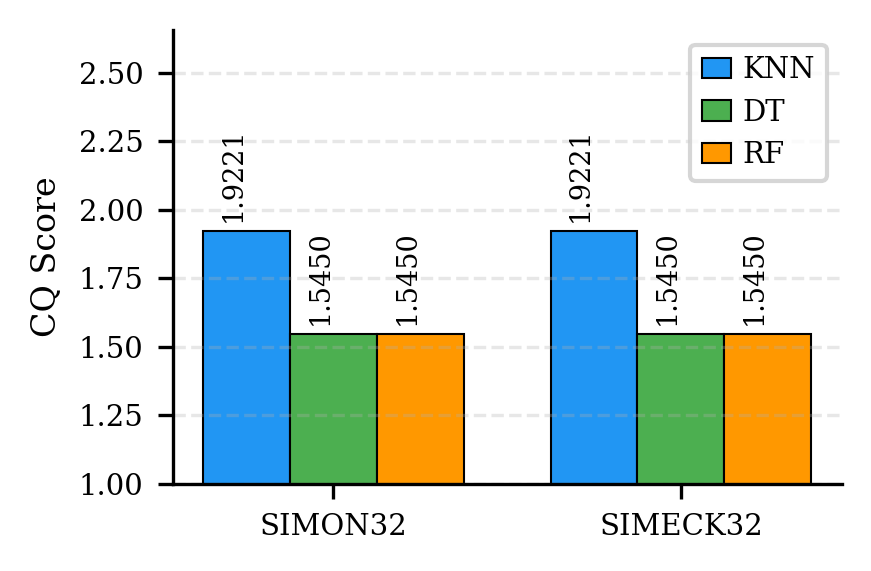}
        \caption{Cluster quality.}
        \label{fig:cluster_quality}
    \end{subfigure}
    \hfill
    \begin{subfigure}[t]{0.48\columnwidth}
        \centering
        \includegraphics[width=\linewidth]{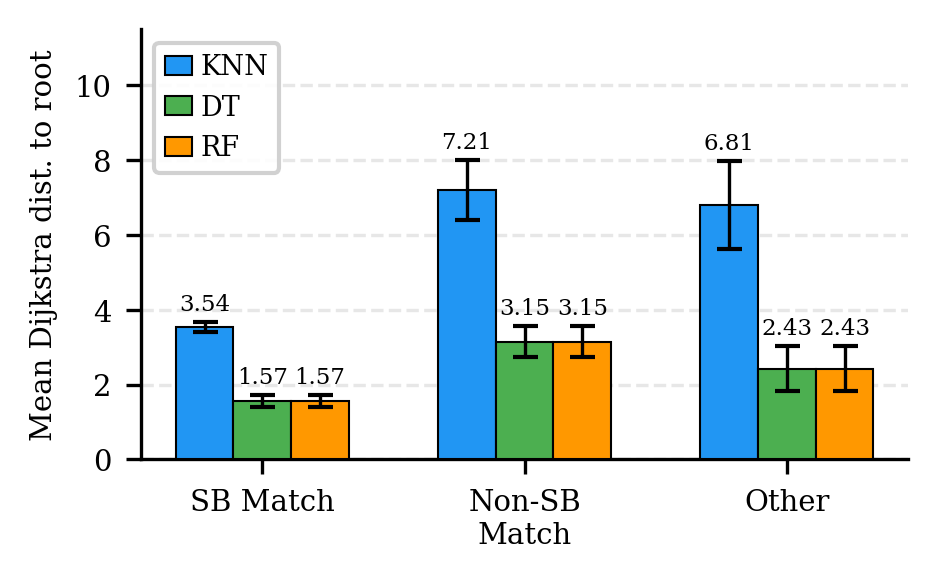}
        \caption{Dijkstra distance to root by group.}
        \label{fig:cluster_proximity}
    \end{subfigure}
    \caption{Clustering proximity and quality.}
    \label{fig:cluster_and_regression}
\end{figure}

As illustrated in Figure \ref{fig:cluster_quality}, KNN achieves the highest CQ score of $1.9221$, meaning other nodes are positioned further from the root than high-priority single-bit differentials. DT and RF produce an identical CQ score of $1.5450$. This result is counterintuitive given that DT and RF are explicitly optimised for differential priority, while KNN is not. Indeed, KNN's neighbourhood structure produces stronger cluster separation in relative terms, albeit at larger absolute distances, as shown in Figure \ref{fig:cluster_proximity}. The tighter intra-cohesion of KNN's high-priority group further confirms that KNN clusters these differentials more compactly around the root in relative graph-distance terms.
Meanwhile, DT and RF produce shorter absolute distances throughout, high-priority nodes sit at a mean distance of $1.5731$ from the root compared to $3.5431$ for KNN, reflecting the sparser, more directed edge structure of the CART-guided graphs. The tighter absolute topology of DT and RF may be preferable in practice, where minimising $E_{hop}$ is the primary objective, while KNN's stronger relative separation may be more informative for differential identification tasks where distinguishing high-priority from non-priority nodes is the goal.
The best differential trail heuristic search results derived from the cluster analysis for all SIMON and SIMECK variants are provided at GitHub \footnote{GitHub URL: https://github.com/johncook1979/Graph-Cluster-Trails-SIMON-SIMECK}, with trails for the largest variants extending beyond previously reported bounds at significantly reduced computational costs \cite{dwivedi2023security, cook2026raid}.

\section{Discussion}\label{Sec:Disc}

This work shows that ML algorithms can effectively identify and structurally separate high-probability differential clusters in GRL, directly addressing the RQ. By enhancing the pDDT with four engineered features, we reveal hidden structural properties, allowing all three models to generate graphs that focus on the stable-state root and significant differentials. The consistent results across both LCAs indicate a response to shared structural properties rather than specific algorithms. Notably, all three models achieve a precision of 1.0 for the optimal $\hat{\Delta} = 1$ and $\Delta_{X_w} = 1$, ensuring no misclassification of differentials. Any node marked as high-priority is guaranteed to be correct, yielding actionable cryptographic insights. The spatial clustering of these nodes visually confirms the differential clustering effect, representing the first such visualisation within a GRL framework.

KNN demonstrates the best cluster separation (CQ = 1.9221, compared to 1.545 for DT and RF) and the highest F1 score ($0.7075$), along with approximately 5 times faster graph construction. It achieves equivalent differential isolation in $\leq 2.4$ seconds on the same hardware as noted in previous studies, showing significant speedup without sacrificing efficiency. While KNN's neighbourhood structure results in tighter intra-cluster cohesion, DT and RF excel in regression accuracy, producing sparser graphs with shorter Dijkstra distances, making them ideal for minimising hops. Overall, KNN is preferable for differential cluster identification and efficiency, while DT and RF are better for compact, path-optimised graphs.

\section{Conclusion and Future Research}\label{Sec:Concl}

This paper presents a GRL framework for the differential cryptanalysis of LCAs by enhancing pDDT differentials with four engineered features, revealing previously hidden structural information for supervised ML models. The resulting features informed three graph-construction algorithms, KNN, DT, and RF, that successfully identified stable-state, high-probability differentials for SIMON$32$ and SIMECK$32$ without false positives. The clustering of differentials in the graph embeddings represents a novel visualisation of the differential clustering effect for both algorithms. The model comparisons highlight a trade-off between strong cluster separation and graph construction time, with KNN showing better separation but at a higher computational cost for DT and RF. The framework appears generalisable across both SIMON$32$ and SIMECK$32$, and future work may expand its evaluation to additional LCAs and larger word sizes, while exploring ways to utilise the identified differential clusters to reduce attack round counts.

\bibliographystyle{splncs04}
\bibliography{references}  






\end{document}